\documentclass[11pt]{article}
\usepackage{nips13submit_e,times}
\usepackage{url}
\usepackage{latexsym}
\usepackage{color}
\usepackage{amsmath}
\usepackage{amssymb}
\usepackage{algorithm2e}
\usepackage{booktabs}
\usepackage{multirow}
\usepackage{tikz}
\usetikzlibrary{trees}
\usepackage{graphics}
\usepackage{wrapfig}

\usepackage{attachfile}

\usepackage{caption} 
\usepackage{titlesec} 
\titlespacing{\section}{0pt}{0pt plus 2pt minus 2pt}{0pt plus 2pt minus 2pt}

\newenvironment{itemize*}%
  {\begin{itemize}%
    \setlength{\itemsep}{0em}%
    \setlength{\parskip}{.5em}}%
  {\end{itemize}}
\newenvironment{enumerate*}%
  {\begin{enumerate}%
    \setlength{\itemsep}{0em}%
    \setlength{\parskip}{.5em}}%
  {\end{enumerate}}
\newenvironment{description*}%
  {\begin{description}%
    \setlength{\itemsep}{0em}%
    \setlength{\parskip}{.5em}}%
  {\end{description}}

\title{Evaluating topic coherence measures}

 \author{F. Rosner\thanks{These authors contributed equally to this work.\newline This work has been presented at the \emph{Topic Models: Computation, Application and Evaluation} workshop at the \emph{Neural Information Processing Systems} conference 2013.}, A. Hinneburg$^*$ \\
   Martin-Luther-University \\
   Halle-Wittenberg, Germany \\
  \And
   M. R\"oder$^*$ \\
    Leipzig University and\\
    Unister GmbH, Germany
    \And
   M. Nettling, A. Both\\
    Unister GmbH, Germany\\
 }

\nipsfinalcopy 

\begin{document}

\abovedisplayskip=10pt 
\belowdisplayskip=10pt 
\abovedisplayshortskip=0pt 
\belowdisplayshortskip=5pt 

\maketitle

\vspace{-9pt}
\begin{abstract}
Topic models extract representative word sets---called topics---from word counts in documents without requiring any semantic annotations. 
Topics are not guaranteed to be well interpretable, therefore, coherence measures have been proposed to distinguish between good and bad topics.
Studies of topic coherence so far are limited to measures that score pairs of individual words.
For the first time, we include coherence measures from scientific philosophy that score pairs of more complex word subsets and apply them to topic scoring.
%
\end{abstract}

\section{Introduction}

Topic model inference searches for a representation of word count distributions of documents as combination of some topic distributions based on word counts.
It is not considered how well derived topics (i.e.,~the top words) can be interpreted by humans.
Recently, coherence measures \cite{Mimno:2011:OSC:2145432.2145462,Newman:2010:AET:1857999.1858011} have been proposed to distinguish between good and bad topics based on top words with respect to interpretability.
Topic models have been combined with coherence measures by introducing specific priors on topic distributions \cite{Mimno:2011:OSC:2145432.2145462,NIPS2011_0366}.

It is interesting to note that all coherence measures evaluated so far take a set of words as input and compute a sum of scores over pairs of words from the input set \cite{Stevens:2012:ETC:2390948.2391052}.
This falls short for examples like \emph{\{bow, tie, match, deck\}} where the word pairs have semantic relations (e.g.,~\emph{\{bow, deck\}} as parts of a ship or  \emph{\{bow, tie\}} as terms of music or clothing) but the set as a whole is not interpretable.
However, a coherence measure based on word pairs would assign a good score.
In scientific philosophy measures have been proposed that compare pairs of more complex word subsets instead of just word pairs.
The main contribution of this paper is to compare coherence measures of different complexity with human ratings.
Furthermore, we include in our study not just word sets generated from topics found by some topic model, but we examine word sets derived by direct optimization of a coherence measure.
This tests whether a coherence measure specifies a useful optimization goal on its own terms.

In Section \ref{sec:coherenceMeasures}, we briefly review coherence measures proposed  in different scientific communities:  NLP, computational linguistics and scientific philosophy.
Section \ref{sec:evaluation} shows the setup and results of our evaluation study and in Section \ref{sec:conclusion} we discuss the results and conclude the paper.

\section{Coherence Measures}
\label{sec:coherenceMeasures}

Coherence measures have been proposed in the NLP community to evaluate topics constructed by some topic model.
In a more general setting, coherence measures have been discussed in scientific philosophy as a formalism to quantify the hanging and fitting together of information pieces \cite{springerlink:10.1007/s11229-006-9131-z}.


Topic coherence has been proposed as an intrinsic evaluation method for topic models \cite{Newman:2010:AET:1857999.1858011,Newman:2010:ETM:1816123.1816156}.
It is defined as average or median of pairwise word similarities formed by top words of a given topic.
Word similarity is grounded on external data not used during topic modeling \cite{NIPS2011_0366}.
The \textbf{UCI-coherence} uses point wise mutual information (PMI) and word cooccurrence counts collected from Wikipedia based on a boolean window model\footnote{We removed stopwords and used a sliding window size of ten words.}.
This word similarity measure induces orderings from bad to good topics that come closest to human coherence judgements.

In \cite{Mimno:2011:BCT:2145432.2145459} inferred posterior distributions of topics are visually analyzed how well they fit the real observations.
However, no coherence measure is proposed to automattically judge interpretability of word sets. 
The coherence measure proposed in \cite{Mimno:2011:OSC:2145432.2145462} is also based on cooccurrences of word pairs. 
Given an ordered list of words $T = \langle w_1,\ldots,w_n \rangle$ the \textbf{UMass-coherence} is defined as
\begin{equation}
\mathcal{C}_{\text{UMass}}(T) = \sum_{m=2}^{M}\sum_{l=1}^{m-1} \log \frac{p(w_m, w_l) + \frac{1}{D}}{p(w_l)}
\end{equation}
A boolean document model is assumed to estimate word probabilities $p$, i.e.,~$p(w_m,w_l)$ is the ratio of number of documents containing both words $w_m, w_l$ and the total number of documents in the corpus $D$.
The smoothing count $1 / D$ is added to avoid calculating the logarithm of zero.

Formalized coherence measures have been proposed in scientific philosophy \cite{bovens_hartmann_coherence_2003,springerlink:10.1007/s11229-011-0003-9,springerlink:10.1007/s11229-010-9856-6,SJP:SJP89}. 
The framework proposed in \cite{springerlink:10.1007/s11229-006-9131-z} unifies several concepts and introduces general qualitative and quantitative coherence measures.
We describe the general framework in the context of word sets. 
Let $W = \{w_1,\ldots,w_n\}$ be a set of words and $W' \subseteq W$.
According to boolean document model, let $p(W')$ be the ratio of the number of documents containing all words of $W'$ divided by the number of documents $D$.
The following three different qualitative coherence notions check whether certain subsets of $W$ increase the conditional probability of other subsets of $W$.  
This is formalized by defining pairs of word subsets $(W',W^*)$ with $W',W^*\subseteq W$ and requiring that $W^*$ supports $W'$, i.e.,~$p(W' | W^*) > p(W')$ holds for all of the required pairs. 
\textbf{One-all coherence} requires that each word $w$ is supported by the complement $W \setminus \{w\}$.
The more complex \textbf{one-any coherence} checks that each word $w$ is supported by all subsets $W^* \subseteq W \setminus \{w\}$. 
The most restrictive \textbf{any-any coherence} requires that each possible, non-empty subset $W' \subset W$ is supported by all other, non-overlapping subsets $W^* \subseteq W\setminus W'$.
Formally, the sets of word set pairs are defined as:
\begin{align*}
S_\text{one-all}(W) &= \big\{(W',W^*)\colon W' = \{w\}, w\in W, W^* = W \setminus W'\big\}\\
S_\text{one-any}(W) &= \big\{(W',W^*)\colon W' = \{w\},w\in W, W^* \subseteq W \setminus W'\big\}\\
S_\text{any-any}(W) &= \big\{(W',W^*)\colon W' \subset W, W^* \subseteq W\setminus W'\big\}
\end{align*}
Quantitative coherence measures $\mathcal C_{d,x}(W)$ are derived by averaging some confirmation measure $d(\cdot,\cdot)$, which quantifies how strong $W^*$ supports $W'$, over all pairs of subsets $S_x(W),\ x \in \{\text{one-all},\ \text{one-any},\ \text{any-any}\}$, depending on the coherence type $x$.
Following \cite{springerlink:10.1007/s11229-006-9131-z}, we use difference measure (a.k.a. interest in association rule mining literature) as confirmation measure. 
Given two non-overlapping subsets $W', W^* \subseteq W$ this measure is defined as:
\begin{equation*}
d(W',W^*) = p(W' | W^*) - p(W')
\end{equation*}
To avoid artifacts, conditional probabilities are neglected for quantitative coherence, when they are computed on a very small subset of the corpus, i.e.,~the condition specifies a subset of ten or less documents.
Note that the framework is very flexible. Both, probability estimation and confirmation measure could be substituted by boolean window model or PMI as in \cite{Newman:2010:AET:1857999.1858011} respectively.

Run time complexities of all proposed coherence measures depend on the number of word set pairs. 
One-all coherence is linear, UMass and UCI are quadratic and one-any as well as any-any coherence are exponential in size of word set $W$.
Despite, the latter two coherences have exponential running times, we assume that their application in practice is possible.
Many techniques from mining frequent item sets may be borrowed that exploit sparsity in text data.
However, a detailed discussion is beyond the scope of this paper.

\section{Evaluation}
\label{sec:evaluation}

We evaluated coherence measures from Section \ref{sec:coherenceMeasures} in three experiments on word sets generated from English and German Wikipedia articles.
The two corpora used consist of articles containing the terms ``movie'' and the German translation ``film'' respectively to ensure that the human raters are famliar with the subject. 
Preprocessing removed redirection and disambiguation pages, portal and category articles as well as articles about single years.
Only nouns have been retained in lemmatized form, except common first names. 
Furthermore, frequent ($\geq 60\%$) and rare nouns ($\leq 1\%$ of documents) have been removed.
The resulting English (German) corpus has 125.410 (71.134) documents, 21.370.741 (6.958.206) tokens and 2.888 (1.885) unique terms.\footnote{corpora and results are available at \url{http://topics.labs.bluekiwi.de/data/nips2013}}

\begin{table}
	\begin{tabular}{lp{0.00125cm}rrrp{0.00125cm}rrr}
	 \toprule
	 \multirow{2}{*}{Coherence} && \multicolumn{3}{c}{German} && \multicolumn{3}{c}{English}\\
	  && good (\%) & neutral (\%) & bad (\%) && good (\%) & neutral (\%) & bad (\%) \\
	 \cmidrule{1-1} \cmidrule{3-5} \cmidrule{7-9}
	Any-any && \textbf{66} & 27 & \textbf{7} &&  52 & 41 & 7 \\
	One-any && 63 & 30 & \textbf{7} && \textbf{59} & 35 & \textbf{6} \\
	One-all && 44 & 36 & 20 && \textbf{59} & 31 & 10 \\
	UMass && 2 & 37 & 61 && 0 & 22 & 78 \\
	Random && 0 & 12 & 88 && 0 & 0 & 100\\
	 \bottomrule
	\end{tabular}
	\centering
	\caption{Human ratings of word sets (Exp.~I).}
	\label{tab:ratingsCoherence}
\end{table}

The first experiment evaluates whether a coherence measure specifies a useful optimization goal on its own terms.
The ability of the coherence measures to mimic human judgements is tested in the second experiment.
The third experiment investigates the applicability of the coherences to topic modeling.

\textbf{Experiment I}.
We generated word sets by directly optimizing coherence using heuristic beam search \cite{Norvig1992}.
A beam search is initialized with a word set comprising of a single word.
Then the algorithm evaluates all possible extensions by another word.
It keeps the $k$ word sets which have the largest coherences.
Those are recursively extended by the same principle until a predefined word set length $l$ is reached.
Thus, $k^{l-1}$ word sets are generated for a given initial word. 
For both corpora, the 20 top TF-IDF terms have been selected as initial terms.
Given some coherence measure and choosing beamwidth $k=3$ and length of word sets $l=5$, $3^{5-1}=81$ word sets are generated for each initial word, thus, $81*20=1620$ words sets in total.
Except UCI coherence -- which does not rely on the given corpus -- for every coherence we randomly sampled 100 of these word sets for human rating. 
Additionally we created 100 word sets randomly as baseline, thus, 500 word sets in total.
Each word set was rated by at least three different human volunteers\footnote{All 19 volunteers that participated are German native speakers and fluent in English.} regarding its interpretability as either good (all five words are related to each other), neutral (three or four words are related) or bad (at most two words are related). 
The kappa statistics \cite{landis1977measurement} about the agreements among the volunteers are $\kappa=0.595$ and $\kappa=0.49$ for German and English data respectively.
The smaller kappa for English word sets might be due to the volunteers' lesser faculty of speech in the foreign language than in their mother tongue.
Table \ref{tab:ratingsCoherence} shows the percentage of word sets regarding the ratings of the majority per coherence measure used for construction.

\begin{table}
	\begin{tabular}{lp{0.125pt}rrp{0.125pt}rr}
	\toprule
	\multirow{2}{*}{Coherence} && \multicolumn{2}{c}{Coh. Word Sets (Exp.~II)} && \multicolumn{2}{c}{LDA Topics (Exp.~III)}\\
  		&& English & German &&  English & German\\
	\cmidrule{1-1} \cmidrule{3-4}  \cmidrule{6-7}
	Any-any && 0.557 & 0.568 && 0.239 & \textbf{0.379} \\
	One-any && \textbf{0.592} & \textbf{0.583} && \textbf{0.242} & 0.376 \\
	One-all && 0.561 & 0.578 && 0.215 & 0.337 \\
	UMass && 0.074 & 0.279 && 0.066 & 0.243 \\
	UCI && 0.224 & 0.380 && 0.219 & 0.371 \\
	\bottomrule
	\end{tabular}
	\centering
	\caption{Kendall's tau rank correlations of coherences and average human ratings (Exp.~II \& III).}
	\label{tab:kendalTauCorrelations}
\end{table}

\textbf{Experiment II}.
For each of the 500 word sets of the first experiment, all five coherence measures are derived.
In Table \ref{tab:kendalTauCorrelations}, second and third columns show the rank correlations between the orderings of the topics by average score of human ratings and the respective coherence measure.

\textbf{Experiment III}.
For each corpus we generated 100 topics using Latent Dirichlet Allocation (LDA) \cite{blei2003latent}\footnote{We used Mallet\cite{mallet2002} with $\alpha=0.5$ and $\beta=0.01$.}.
Each topic has been rated by human volunteers ($\kappa=0.45$ for German and $\kappa=0.29$ for Englisch topics). 
In Table \ref{tab:kendalTauCorrelations}, columns four and five report the rank correlations between the orderings of the topics by average score of human ratings and the respective coherence measure.



\section{Discussion and Conclusion}
\label{sec:conclusion}

The results of the first experiment show that if we are using the one-any, any-any and one-all coherences directly for optimization they are leading to meaningful word sets.
The second experiment shows that these coherence measures are able to outperform the UCI coherence as well as the UMass coherence on these generated word sets.
For evaluating LDA topics any-any and one-any coherences perform slightly better than the UCI coherence.
The correlation of the UMass coherence and the human ratings is not as high as for the other coherences.

Our results clearly show that comparing just word pairs via conformation measures can lead to poor performing coherence measures.
This indicates that evaluating word pairs is not enough to mimic human ratings.

Our results might give rise to the development of new priors for topic models.
However, directly optimizing coherence may lead to other meaningful word sets that maybe missed by topic models.
Therefore, it is worth to explore multi-criteria optimization for learning topic models instead of combining coherence measures as prior distributions with topic model inference. Additional future work should focus on exploring the characteristics of the different coherences.
This should include the different conformation measures, the document model used and possible requirements which are imposed on the corpus by the coherences.

\section*{Acknowledgments}

\begin{wrapfigure}[2]{r}{4.2cm}
 \vspace{-6mm}
 \includegraphics[scale=0.333]{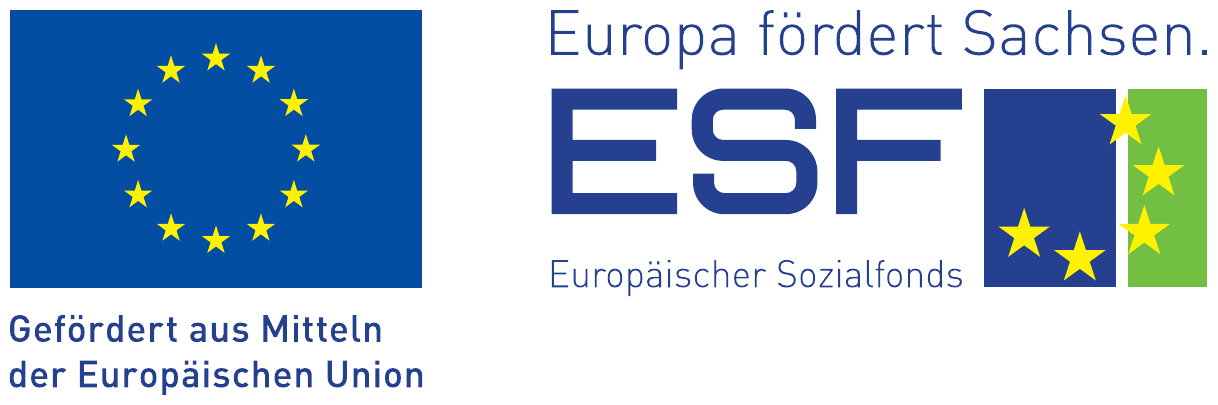}
\end{wrapfigure}
We thank all volunteers who participated in the evaluation.
Parts of this work were supported by the ESF and the Free State of Saxony.

{\small

}

\end{document}